\documentclass{article} % For LaTeX2e

\usepackage{nips15submit_e,times}
\usepackage[colorlinks=True,citecolor=blue,linkcolor=blue]{hyperref}
\usepackage{url}
\usepackage{amsmath}
\usepackage{amsfonts}

\usepackage{multirow}

\usepackage[pdftex]{graphicx}
\graphicspath{{img/}{extras/}}
\DeclareGraphicsExtensions{.pdf,.jpeg,.png}
\usepackage{rotating}

\title{Attention-Based Models for Speech Recognition} 

\author{
Jan Chorowski \\
University of Wroc\l{}aw, Poland\\
\texttt{jan.chorowski@ii.uni.wroc.pl} \\
\And
Dzmitry Bahdanau \\
Jacobs University Bremen, Germany
\And
Dmitriy  Serdyuk \\
Universit\'{e} de Montr\'{e}al
\And
Kyunghyun Cho\\
Universit\'{e} de Montr\'{e}al
\And
Yoshua Bengio \\
Universit\'{e} de Montr\'{e}al \\
CIFAR Senior Fellow
}

\nipsfinalcopy % Uncomment for camera-ready version

\begin{document}

\maketitle

\begin{abstract}
% TODO: insert final correct numbers    
Recurrent sequence generators conditioned on input data through an attention
mechanism have recently shown very good performance on a range of tasks
including machine translation, handwriting synthesis
\cite{graves_generating_2013,bahdanau_neural_2014} and image caption generation
\cite{xu_show_2015}. We extend the attention-mechanism with features needed for speech
recognition. We show that while an adaptation of the model used for machine
translation in \cite{bahdanau_neural_2014} reaches a competitive 18.7\% phoneme
error rate (PER) on the TIMIT phoneme recognition task, it can only be applied to utterances which are
roughly as long as the ones it was trained on. We offer a qualitative
explanation of this failure 
and propose a novel and generic method of adding
location-awareness to the attention mechanism to alleviate this issue. The new
method yields a model that is robust to long inputs and achieves 18\% PER in single
utterances and 20\% in 10-times longer (repeated) utterances.  Finally, we
propose a change to the attention mechanism that prevents it from concentrating
too much on single frames, which further reduces PER to 17.6\% level. 
\end{abstract}

\section{Introduction}

% Paragraph 1: introduction attention mechanisms
Recently, attention-based recurrent networks have been successfully applied to a
wide variety of tasks, such as handwriting
synthesis~\cite{graves_generating_2013}, machine
translation~\cite{bahdanau_neural_2014}, image caption
generation~\cite{xu_show_2015} and  visual object
classification~\cite{mnih_2014}.\footnote{%Stop space
  An early version of this work was presented at the NIPS 2014 Deep Learning
  Workshop \cite{chorowski_2014}.
}
Such models iteratively process their input by selecting relevant content at
every step. This basic idea significantly extends the applicability range of
end-to-end training methods, for instance, making it possible to construct
networks with external memory~\cite{graves_2014,weston_2014}.

% Paragraph 3: why speech recognition with attention is an 
% an interesting machine learning problem.
% IMPORTANT, THE STORY
We introduce extensions to attention-based recurrent networks
that make them applicable to
speech recognition. Learning to recognize speech can be viewed as
learning to generate a sequence (transcription) given another sequence (speech).
From this perspective it is similar to machine translation and handwriting
synthesis tasks, for which attention-based methods have been found suitable
\cite{bahdanau_neural_2014,graves_generating_2013}.  However, compared to
machine translation, speech recognition principally differs by
requesting much longer input
sequences (thousands of frames instead of dozens of words), which introduces a
challenge of distinguishing similar speech fragments\footnote{Explained in
more detail in Sec.~\ref{subsec:framework}.} in a single utterance.
% as
% well as issues of long-term dependencies~\cite{Bengio-trnn93-small} and the
% presence of two different scales in the input and output sequences. 
% DIMA:  both these issues are revelent for machine and
% translation and handwriting recognition. Let's
% concentrate on key points.
It is
also different from handwriting synthesis, since the input sequence is much
noisier and does not have as clear structure. For these reasons speech
recognition is an interesting testbed for developing new attention-based
architectures capable of processing long and noisy inputs.

% Paragraph 2: introduction to speech recognition
Application of attention-based models to speech recognition is also an
important step toward building fully end-to-end trainable speech
recognition systems, which is an active area of research. The dominant
approach is still based
on hybrid systems consisting of a deep neural acoustic model, a triphone HMM
model and an n-gram language
model~\cite{gales_application_2007,hinton_deep_2012}. This requires dictionaries
of hand-crafted pronunciation and phoneme lexicons, and a multi-stage training
procedure to make the components work together. Excellent results by an HMM-less
recognizer have recently been reported, with the system consisting of a
CTC-trained neural network and a language model~\cite{hannun2014_deepspeech}.
Still, the language model was added only at the last stage in that work, thus
leaving open a question of how much an acoustic model can benefit from being
aware of a language model during training.

% Paragraph 4: what we actually do
In this paper, we evaluate attention-based models on a
phoneme recognition task using the widely-used TIMIT
dataset. At each time step in generating an output sequence (phonemes),
an attention mechanism selects or weighs the signals produced
by a trained feature extraction mechanism at potentially all of the time steps 
in the input sequence (speech frames). The weighted feature vector then
helps to condition the generation of the next element of the output sequence.
Since the utterances in this dataset are rather
short (mostly under 5 seconds), we measure the
ability of the considered models in recognizing much longer
utterances which were created by artificially concatenating
the existing utterances.

% Paragraph 5: first important finding is the deceptively
% good performance of the baseline
We start with a model proposed in
\cite{bahdanau_neural_2014} for the machine translation task
as the baseline. This model seems entirely vulnerable to
the issue of similar speech fragments 
but despite our
expectations it was competitive on the original test set, reaching 18.7\% phoneme
error rate (PER). However, its performance
degraded quickly with longer, concatenated utterances. We
provide evidence that this model adapted to track the absolute
location in the input sequence of the content it is
recognizing, a strategy feasible for short utterances from the
original test set but inherently unscalable.

% Paragraph 6: the main contribution
In order to circumvent this undesired behavior, in this
paper, we propose to modify the attention mechanism such
that it explicitly takes into account both (a) the location of the
focus from the previous step, as in~\cite{graves_2014} and (b)
the features of the input sequence, as in~\cite{bahdanau_neural_2014}. 
This is achieved by adding as inputs to the attention mechanism
auxiliary {\it convolutional features} which are extracted 
by convolving the attention
weights from the previous step with trainable filters.  We show that a model
with such convolutional features performs significantly
better on the considered task (18.0\% PER). More
importantly, the model with convolutional features robustly
recognized utterances many times longer than the ones
from the training set, always staying below 20\% PER.

% Paragraph 7: smoothing?? windowing?? should we add it?

Therefore, the contribution of this work is three-fold. For one, we present a
novel purely neural speech recognition architecture based on an attention
mechanism, whose performance is comparable to that of the conventional
approaches on the TIMIT dataset.  Moreover, we propose a generic method of
adding location awareness to the attention mechanism. Finally, we introduce a
modification of the attention mechanism to avoid concentrating the attention on
a single frame, and thus avoid obtaining less ``effective training examples'',
bringing the PER down to 17.6\%.

\section{Attention-Based Model for Speech Recognition}

\subsection{General Framework}
\label{subsec:framework}

An attention-based recurrent sequence generator (ARSG) is a recurrent neural
network that stochastically generates an output sequence $(y_1, \dots, y_T)$
from an input $x$.  In practice, $x$ is often processed by an {\it encoder}
which outputs a sequential input representation $h=(h_1,\ldots,h_L)$ more
suitable for the attention mechanism to work with.

In the context of this work, the output $y$ is a sequence of phonemes, and the
input $x=(x_1, \ldots, x_{L'})$ is a sequence of feature vectors. Each feature
vector is extracted from a small overlapping window of audio frames. The encoder
is implemented as a deep bidirectional recurrent network (BiRNN), to form a
sequential representation $h$ of length $L=L'$.

\begin{figure}
    \centering
    %with .65\textwidth the font size roughly matches text
    \hfill
    \begin{minipage}{0.6\textwidth}
        \centering
        \includegraphics[width=0.95\columnwidth]{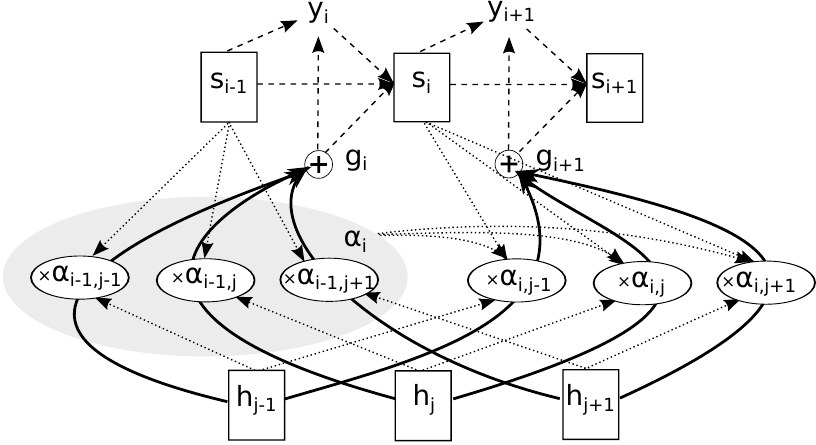}
    \end{minipage}
    \begin{minipage}{0.38\textwidth}
        \caption{
            Two steps of the proposed attention-based recurrent sequence
            generator (ARSG) with a hybrid attention mechanism (computing $\alpha$), based on
            both content ($h$) and location (previous $\alpha$) information.
            The dotted lines correspond to Eq.~\eqref{eq:attention}, thick solid
            lines to Eq.~\eqref{eq:glimpse} and dashed lines to
            Eqs.~\eqref{eq:generate}--\eqref{eq:recurrency}.
        }
        \label{fig:model}
    \end{minipage}
    \hfill
    %\includegraphics[width=.5\textwidth]{model.pdf}
    % As per the style file: the extra space here is intentional
    % \vspace{-.8cm}
    %\caption{Two steps of an ARSG with a hybrid attention
    %    mechanism. The dotted lines correspond to
    %    equation \eqref{eq:attention}, thick solid lines to \eqref{eq:glimpse} and dashed lines
    %    to \eqref{eq:generate} and \eqref{eq:recurrency}.
    %}
    %\label{fig:model}

  \vspace{-4mm}
\end{figure}

At the $i$-th step an ARSG generates an output $y_i$ by focusing on the relevant
elements of $h$:
\begin{align}
    \alpha_i = Attend(s_{i-1},\alpha_{i-1},h) 
    \label{eq:attention}
    \\
    g_i = \sum\limits_{j=1}^L \alpha_{i,j} h_j 
    \label{eq:glimpse}
    \\
    y_i \sim Generate(s_{i-1}, g_i),
    \label{eq:generate}
\end{align}
where $s_{i-1}$ is the $(i-1)$-th state of the recurrent neural network to which
we refer as the {\it generator}, $\alpha_i \in \mathbb{R}^L$ is a vector of the
{\it attention weights}, also often called the
alignment~\cite{bahdanau_neural_2014}. Using the terminology from
\cite{mnih_2014}, we call $g_i$ a {\it glimpse}. The step is completed by
computing a new generator state:
\begin{align}
s_i = Recurrency(s_{i-1}, g_i, y_i)
\label{eq:recurrency}
\end{align}
Long short-term memory units (LSTM, \cite{hochreiter_1997}) and gated recurrent
units (GRU, \cite{cho_2014}) are typically used as a recurrent activation, to
which we refer as a {\it recurrency}.  The process is graphically illustrated in
Fig.~\ref{fig:model}.

Inspired by \cite{graves_2014}  we distinguish between location-based,
content-based and hybrid attention mechanisms. $Attend$ in
Eq.~\eqref{eq:attention} describes the most generic, hybrid attention. If the
term $\alpha_{i-1}$ is dropped from $Attend$ arguments, i.e.,
$\alpha_i=Attend(s_{i-1},h)$,
we call it content-based (see, e.g., \cite{bahdanau_neural_2014} or
\cite{xu_show_2015}). In this case, $Attend$ is often implemented by scoring
each element in $h$ separately and normalizing the scores:
\begin{align}
  \label{eq:pure_cb_attention}
    e_{i,j}=Score(s_{i-1}, h_j), \\ 
  \label{eq:softmax_normalization}
    \alpha_{i,j} = 
        \exp(e_{i,j}) \left/
        \sum\limits_{j=1}^L \exp(e_{i,j}) \right..
\end{align}    

%
% JAN: While I understand this point, I don't like it. The net has finite
% capacity (because we assume finite float precision and fixed
% number of units). Then it must fail for sequences which can be
% arbitrarily long.
%
% OTOH, If you assume unbounded capacity of the representation, then
% how do you learn it?
%
% DIMA: Agreed, we should rephrase it later.
%
% Cho: I rephrased it a bit, but now I feel that what I wrote is too dramatic.

The main limitation of such scheme is that identical or very similar elements of
$h$ are scored equally regardless of their position in the
sequence. This is the issue of ``similar speech fragments'' raised above.
Often this issue is partially alleviated by an encoder such
as e.g. a BiRNN~\cite{bahdanau_neural_2014} or a deep convolutional
network~\cite{xu_show_2015} that encode contextual
information into every element of $h$ . However, capacity of
$h$ elements is always limited, and thus disambiguation by
context is only possible to a limited extent.

%In
%practice $h$ elements often contain a lot of contextual information, being e.g.
%states of a BiRNN or feature vectors extracted by a deep convolutional network
%\cite{xu_show_2015}, which partially alleviates the issue. However, their
%capacity is still limited, which imposes certain upper bound on the the maximum
%size of the input $x$ that can be handled.

Alternatively, a location-based attention mechanism computes the alignment from
the generator state and the previous alignment only such that
$\alpha_{i} = Attend(s_{i-1}, \alpha_{i-1})$. 
For instance, Graves \cite{graves_generating_2013} used the location-based
attention mechanism using a Gaussian mixture model in his handwriting synthesis
model.  In the case of speech recognition, this type of location-based attention
mechanism would have to predict the distance between consequent 
phonemes using $s_{i-1}$ only, which we expect to be hard due to
large variance of this quantity.

For these limitations associated with both content-based and location-based
mechanisms, we argue that a hybrid attention mechanism is a natural candidate
for speech recognition. Informally, we would like an attention model that uses
the previous alignment $\alpha_{i-1}$ to select a short list of elements from
$h$, from which the content-based attention, in
Eqs.~\eqref{eq:pure_cb_attention}--\eqref{eq:softmax_normalization}, will select
the relevant ones without confusion.

\subsection{Proposed Model: ARSG with Convolutional Features}

We start from the ARSG-based model with the content-based attention mechanism
proposed in \cite{bahdanau_neural_2014}. This model can be described by
Eqs.~\eqref{eq:pure_cb_attention}--\eqref{eq:softmax_normalization}, where 
\begin{align}
e_{i,j} = w^\top \tanh(W s_{i-1} + V h_j + b).
\label{eq:base_attention}
\end{align}
$w$ and $b$ are vectors, $W$ and $V$ are matrices. 

% %TODO: JAN what to do with this 
% We further consider the following location-based additions
% to it. Define $p_i$ the ``expected'' position of the Generator
% at step $i$:
% \begin{align}
%     p_i = \sum\limits_{j=1} \alpha_{i,j} j
% \end{align}
% One natural idea is that the positions to the left of $p_i$ or far
% to the right of $p_i$ are unlikely to be useful for
% generating $y_{i+1}$. We add a \textit{gating mechanism} $g$ to make
% the model capable of learning it:
% \begin{align}
%   \label{eq:gating}
%     g(x) = w_g \tanh(v_g x) \\
%     e_{i,j} = w^T \tanh(W s_{i-1} + V h_j) g(j - p_i)
% \end{align}
% where $w_g$ and $v_g$ are parameter vectors trainable with
% the rest of the model.

% The gating mechanism can be critized for being a shallow
% addition to the attention model, only applied at the latest
% stage. It can also be affected by non-robustness of the
% mean. For instance, if the correct location were given a
% weight of 0.9, and a wrong location 500 steps to the right
% were given a weight of 0.1, the expected position mean $p_i$
% would be 50 position wrong, which would very likely harm the
% decoding procedure.  Finally, the gater $g$ is a rather
% unconventional scalar-to-scalar network whose arguments span
% a very large range. Therefore it required a special
% initialization in order to be trained in our experiments.

We extend this content-based attention mechanism of the original model to be
location-aware by making it take into account the alignment produced at the
previous step. First, we extract $k$ vectors $f_{i,j} \in \mathbb{R}^{k}$ for
every position $j$ of the previous alignment $\alpha_{i-1}$  by convolving it
with a matrix $F \in \mathbb{R}^{k \times r}$:
\begin{align}
    \label{eq:conv_feats}
    f_i = F * \alpha_{i-1}.
\end{align}
These additional vectors $f_{i,j}$ are then used by the scoring mechanism $e_{i,j}$:
\begin{align}
    \label{eq:hybrid_score}
    e_{i,j} = w^\top \tanh(W s_{i-1} + V h_j + U f_{i,j} + b)
\end{align}

\subsection{Score Normalization: Sharpening and Smoothing}
\label{sec:sharpening}

There are three potential issues with the normalization in
Eq.~\eqref{eq:softmax_normalization}. 

First, when the input sequence $h$ is long, the glimpse $g_i$ is likely to
contain noisy information from many irrelevant feature vectors $h_j$, as the
normalized scores $\alpha_{i,j}$ are all positive and sum to $1$. This makes it
difficult for the proposed ARSG to focus clearly on a few relevant frames at
each time $i$. Second, the attention mechanism is required to consider all the
$L$ frames each time it decodes a single output $y_i$ while decoding the output
of length $T$, leading to a computational complexity of $O(LT)$. This may
easily become prohibitively expensive, when input utterances are long
(and issue that is less serious for machine translation, because in that
case the input sequence is made of words, not of 20ms acoustic frames).

The other side of the coin is that the use of {\it softmax} normalization in
Eq.~\eqref{eq:softmax_normalization} prefers to mostly focus on only a single
feature vector $h_j$. This prevents the model from aggregating multiple
top-scored frames to form a glimpse $g_i$.  

\paragraph{Sharpening}

There is a straightforward way to address the first issue of a noisy glimpse by
``sharpening'' the scores $\alpha_{i,j}$. One way to sharpen the weights is to
introduce an {\it inverse temperature} $\beta > 1$ to the softmax function such
that
\[
    a_{i,j}=\exp(\beta e_{i,j})\left/ \sum_{j=1}^{L}\exp(\beta e_{i,j})\right.,
\]
or to keep only the top-$k$ frames according to the scores and re-normalize
them.  These sharpening methods, however, still requires us to compute the score
of every frame each time ($O(LT)$), and they worsen the second issue, of overly narrow 
focus.

We also propose and investigate a {\it windowing} technique.
At each time $i$, the attention mechanism considers only a subsequence
$\tilde{h} = (h_{p_i-w}, \ldots, h_{p_i+w-1})$ of the whole sequence $h$, where 
$w \ll L$ is the predefined window width and $p_i$ is the median of the alignment
$\alpha_{i-1}$. The scores for $h_j \notin \tilde{h}$ are not computed, 
resulting in a lower complexity of $O(L+T)$.
This windowing technique is similar to taking the top-$k$ frames, and similarly,
has the effect of sharpening.
%Last but not least, windowing can be
%considered a location-based addition to the attention
%mechanism.

The proposed sharpening based on windowing can be used both during training and
evaluation. Later, in the experiments, we only consider the case where it is
used during evaluation.

\paragraph{Smoothing}

We observed that the proposed sharpening methods indeed
helped with long utterances. However, all of them, and
especially selecting the frame with the highest score,
negatively affected the model's performance on the standard
development set which mostly consists of short utterances.
This observations let us hypothesize that it is helpful for
the model to aggregate selections from multiple top-scored
frames. In a sense this brings more diversity, i.e., more
effective training examples, to the output part of the model,
as more input locations are considered.
To facilitate this effect, we replace the
unbounded exponential
function of the softmax function in
Eq.~\eqref{eq:softmax_normalization} with the
bounded logistic sigmoid $\sigma$ such that 
\[
    a_{i,j}=\sigma(e_{i,j})\left/\sum_{j=1}^{L}\sigma(e_{i,j})\right..
\]
This has the effect of {\it smoothing} the focus found by the attention
mechanism.

\section{Related Work}

Speech recognizers based on the connectionist temporal classification (CTC,
\cite{graves_2006_connectionist}) and its extension, RNN
Transducer~\cite{graves_2012_sequence}, are the closest to the ARSG model
considered in this paper. They follow earlier work on end-to-end trainable deep
learning over sequences with gradient signals flowing through the alignment
process~\cite{LeCun98-small}. They have been shown to perform well on the
phoneme recognition task~\cite{graves_2013_timit}.  Furthermore, the CTC was
recently found to be able to directly transcribe text from speech without any
intermediate phonetic representation~\cite{graves_2014_towards}.

\begin{figure}[t]
  \centering
  \includegraphics[width=\textwidth]{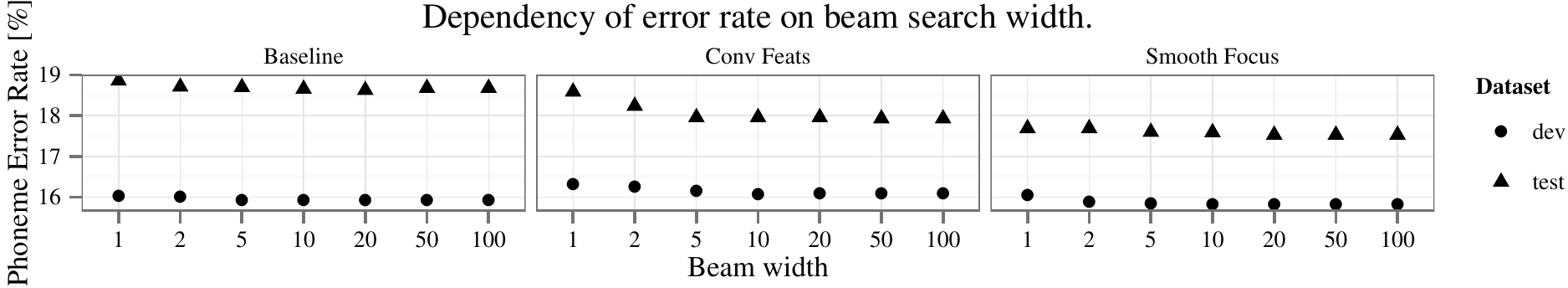}
  % \vspace{-.8cm}
  \caption{Decoding performance w.r.t. the beam size.
      %on the TIMIT test set. 
      For rigorous comparison, if decoding failed to generate
      $\left<\text{eos}\right>$, we considered it wrongly recognized without
      retrying with a larger beams size. The models, especially with smooth focus, 
      perform well even
      with a beam width as small as 1.
      %beam size was
      %not increased even when decoding failed to generate
      %the ``end-of-sequence'' token. In such a case all
      %phonemes of the utterance were considered wrongly
      %recognized.
  }
  \label{fig:beam_dependency}

  \vspace{-4mm}
\end{figure}

The considered ARSG is different from both the CTC and RNN Transducer in two
ways.  First, whereas the attention mechanism deterministically aligns the input
and the output sequences, the CTC and RNN Transducer treat the alignment as a
latent random variable over which MAP (maximum a posteriori) inference is
performed. This deterministic nature of the ARSG's alignment mechanism allows
beam search procedure to be simpler. Furthermore, we empirically observe that a
much smaller beam width can be used with the deterministic mechanism, which
allows faster decoding (see Sec.~\ref{sec:evaluate} and
Fig.~\ref{fig:beam_dependency}).
Second, the alignment mechanism of both the CTC and RNN Transducer is
constrained to be ``monotonic'' to keep marginalization of the alignment
tractable. On the other hand, the proposed attention mechanism can result in
non-monotonic alignment, which makes it suitable for a larger variety of tasks
other than speech recognition.

A hybrid attention model using a convolution operation was also proposed in
\cite{graves_2014} for neural Turing machines (NTM). At each time step, the NTM
computes content-based attention weights which are then convolved with a
predicted shifting distribution. Unlike the NTM's approach, the hybrid mechanism
proposed here lets learning figure out how the content-based and location-based
addressing be combined by a deep, parametric function (see
Eq.~\eqref{eq:hybrid_score}.) 
%Also, it does not need to separately predict the
%shifting distribution.

Sukhbaatar et al. \cite{sukhbaatar_2015} describes a similar hybrid attention
mechanism, where location embeddings are used as input to the attention model.
This approach has an important disadvantage that the model cannot work with an
input sequence longer than those seen during training. Our approach, on the
other hand, works well on sequences many times longer than those seen during
training (see Sec.~\ref{sec:results}.)

\section{Experimental Setup}
\label{sec:setup}

We closely followed the procedure in \cite{graves_2013_timit}. All experiments
were performed on the TIMIT corpus \cite{timit}. We used the train-dev-test
split from the Kaldi \cite{povey_2011} TIMIT s5 recipe. We trained on the
standard 462 speaker set with all SA utterances removed and used the 50 speaker
dev set for early stopping. We tested on the 24 speaker core test set. All
networks were trained on 40 mel-scale filter-bank features together with the
energy in each frame, and first and second temporal differences, yielding in
total 123 features per frame. Each feature was rescaled to have zero mean and
unit variance over the training set. Networks were trained on the full 61-phone
set extended with an extra ``end-of-sequence'' token that was appended to each
target sequence. Similarly, we appended an all-zero frame at the end of each
input sequence to indicate the end of the utterance. Decoding was performed
using the 61+1 phoneme set, while scoring was done on the 39 phoneme set.

\subsection{Training Procedure}

One property of ARSG models is that different subsets of parameters are reused
different number of times; $L$ times for those of the encoder, $LT$ for the
attention weights and $T$ times for all the other parameters of the ARSG.  This
makes the scales of derivatives w.r.t. parameters vary significantly, and we
handle it by using an adaptive learning rate algorithm,
AdaDelta~\cite{zeiler_2012} which has two hyperparameters $\epsilon$ and $\rho$.
All the weight matrices were initialized from a normal Gaussian distribution
with its standard deviation set to $0.01$. Recurrent weights were furthermore
orthogonalized.
%DIMA: reference for orthogonal weights?

As TIMIT is a relatively small dataset, proper regularization is crucial. We used
the adaptive weight noise as a main regularizer~\cite{graves_2011}.  We first
trained our models with a column norm constraint~\cite{hinton_2012} with the
maximum norm $1$
%(all weights incoming to each neuron were clipped to norm 1) 
until the lowest development negative log-likelihood is achieved.\footnote{
    Applying the weight noise from the beginning of training caused severe
    underfitting.
    %At the beginning of training the attention mechanism is not functional.
    %Therefore too much noise is applied to its weights and in consequence it
    %fails to train.
}
During this time, $\epsilon$ and $\rho$ are set to $10^{-8}$ and $0.95$,
respectively.  At this point, we began using the adaptive weight noise, and
scaled down the model complexity cost $L_C$ by a factor of 10, while disabling
the column norm constraints. Once
the new lowest development log-likelihood was reached, we fine-tuned the model
with a smaller $\epsilon=10^{-10}$, until we did
not observe the improvement in the development phoneme error rate (PER) for 100K
weight updates. Batch size 1 was used throughout the
training.

% Cho: I remember Hinton argued for the norm constraint as an alternative to
% weight decay, meaning we probably shouldn't consider them similar.
%were lifted, since adaptive
%weight noise method includes a term which acts similarly to
%weight decay. 

%TODO: what other details are needed there: initialization?
%For more information about the training procedure, see
%Supplementary Material.

% As indicated, in some experiments, the alignments selected by
% the model were limited to a window around the median position of the
% decoder at the previous step.

\subsection{Details of Evaluated Models}
\label{sec:evaluate}

We evaluated the ARSGs with different attention mechanisms.  The encoder was a
3-layer BiRNN with 256 GRU units in each direction, and the activations of the
512 top-layer units were used as the representation $h$.  The generator had a
single recurrent layer of 256 GRU units. 
$Generate$ in Eq.~\eqref{eq:generate} had a hidden layer of 64 maxout units.
The initial states of both the encoder and generator were treated as additional
parameters.

%and were learned for both
%Encoder and Generator.

%Similarly to \cite{bahdanau_neural_2014}, an
%additional layer of 64 maxout units combined the state $s_{i-1}$ and
%the glimpse $g_i$ (c.f. function $Generate$ in
%\eqref{eq:generate}). 

Our baseline model is the one with a purely content-based attention mechanism
(See Eqs.~\eqref{eq:pure_cb_attention}--\eqref{eq:base_attention}.) The scoring
network in Eq.~\eqref{eq:base_attention} had 512 
%$\tanh$ 
hidden units.
%, i.e. matrices $W$ and $V$ in Eq.~\eqref{eq:base_attention} had 512 rows.
%TODO: when removing the gater, remove this
% When used, the gating mechanism $g$ from equation \eqref{eq:gating} 
% had 10 hidden $\tanh$ units. 
The other two models use the convolutional features in Eq.~\eqref{eq:conv_feats}
with $k=10$ and $r=201$. One of them uses the smoothing from
Sec.~\ref{sec:sharpening}.

\paragraph{Decoding Procedure}
A left-to-right beam search over phoneme sequences
was used during decoding \cite{sutskever_sequence_2014}. Beam search was stopped
when the ``end-of-sequence'' token $\left<\text{eos}\right>$ was emitted. We
started with a beam width of 10, increasing it up to 40 when the network failed
to produce $\left<\text{eos}\right>$ with the narrower beam.  As shown in
Fig.~\ref{fig:beam_dependency}, decoding with a wider beam gives little-to-none
benefit.

\section{Results}
\label{sec:results}

\begin{figure}[t]
  \centering
  \includegraphics[width=\textwidth]{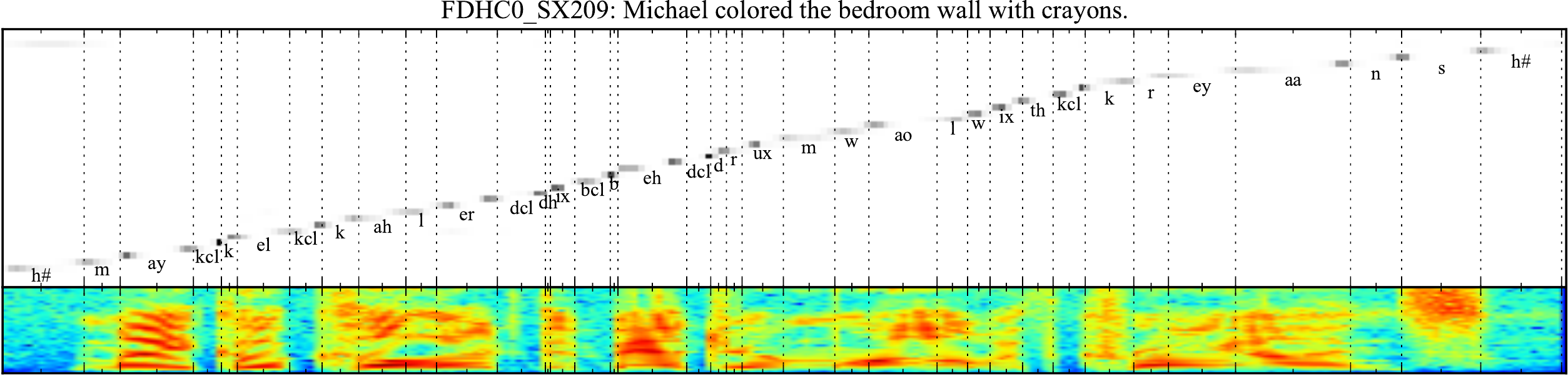}
  % \vspace{-.8cm}
  \caption[Alignments produced by the baseline]{Alignments
      produced by the baseline model. The vertical bars
      indicate ground truth phone location from TIMIT. Each
      row of the upper image indicates frames selected by
      the attention mechanism to emit a phone symbol. 
      %We observe that the 
      % Cho: probably redundant since the text below is already mentioned in the
      % text..
      The network has clearly learned to produce a left-to-right alignment with
      a tendency to look slightly ahead, and does not confuse between the
      repeated ``kcl-k'' phrase. 
      Best viewed in color.
      % TODO: the reviewer can ask what is the difference
      % when smooth focus is used
  }  
  \label{fig:ali_baseline}

  \vspace{-4mm}
\end{figure}

\begin{table}[h]
    \caption{Phoneme error rates (PER). The bold-faced PER corresponds to the
    best error rate with an attention-based recurrent sequence generator (ARSG)
    incorporating convolutional attention features and a smooth focus. }
  \label{tab:results}
  % \vspace{1mm}
  \centering
\begin{tabular}{l|c|c}
  % \hline
  \multicolumn{1}{c|}{\bf Model}  &\multicolumn{1}{c|}{\bf Dev} &\multicolumn{1}{c}{\bf Test} \\ 
  \hline 
  \hline 
  %\multicolumn{3}{c}{Kaldi TIMIT s5 recipe with {\tt basic} scorer} \\
  %\hline 
  % JCH13
  Baseline Model & 15.9\% & 18.7\% \\
  % JCH14
  Baseline + Conv. Features & 16.1\% & 18.0\% \\
  Baseline + Conv. Features + Smooth Focus & 15.8\% & {\bf 17.6\%} \\
  \hline
  % Pre-Transducer
  RNN Transducer 
  \cite{graves_2013_timit} & N/A & 17.7\% \\
  \hline\hline
  HMM over Time and Frequency Convolutional Net
  \cite{toth_2014} & 13.9\% & 16.7\% 
  \end{tabular}

  \vspace{-4mm}
\end{table}

All the models achieved competitive
PERs (see Table~\ref{tab:results}). 
With the convolutional features, we see 3.7\% relative improvement over the
baseline and further 5.9\% with the smoothing. 

%, which increased to 5.4\% when we also
%switched to the smooth-focus attention. 

% Cho: Maybe unnecessary to say, unless the other results (RNN Transducer and
% HMM T-F Conv Net) reported the variance.
%Every experiment was
%repeated only once, thus variance of these results is
%unknown. 

% Cho: Maybe in the conclusion?
%We note, that we also experimented with the gating
%mechanism proposed in the preliminary version of this work
%\cite{chorowski_2014}. We found it however in all aspects
%inferior to the convolutional features and subjected to
%instability of the mean.

To our surprise
%create proper alignments 
%considerations that a content-based attention can not do so 
(see Sec.~\ref{subsec:framework}.),  
the baseline model learned to align properly.
An alignment produced by the baseline model
on a sequence with repeated phonemes (utterance FDHC0\_SX209) is presented in
Fig.~\ref{fig:ali_baseline} which demonstrates that  the baseline model is not
confused by short-range repetitions. We can also see
from the figure that it
prefers to select frames that are near the beginning or even slightly before the
phoneme location provided as a part of the dataset. The alignments produced by
the other models were very similar visually.

\begin{figure}[t]
  \centering
  \vspace{.1cm} %only to visually separate from Table 1
  \includegraphics[width=\textwidth]{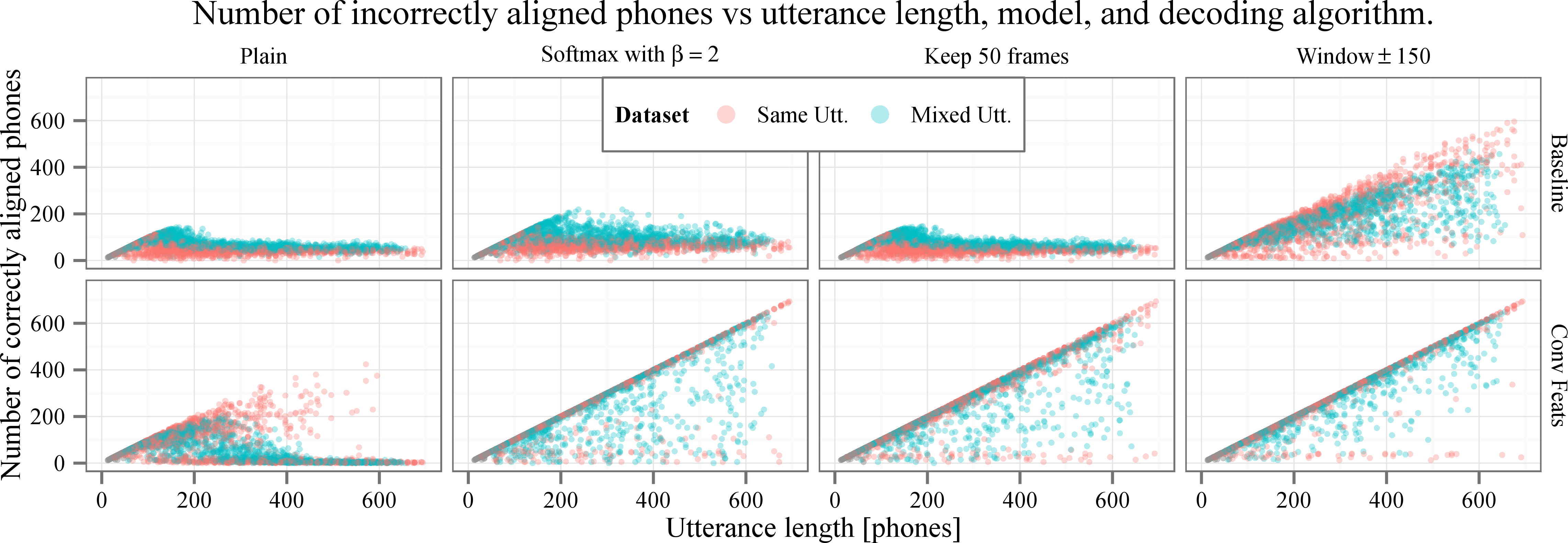}
  %\includegraphics[width=\textwidth]{alis_corrected.pdf}
  %\vspace{-.8cm}
  \caption[Results of force-aligning of long utterances with corrections.]{
    Results of force-aligning the concatenated utterances. Each dot
    represents a single utterance created by either concatenating
    multiple copies of the same utterance, or of different, randomly
    chosen utterances. 
    %We compare the models with either a content-based only
    %or a hybrid attention mechanism.
    We clearly see that the highest robustness is achieved
    when the hybrid attention mechanism is combined with the proposed sharpening
    technique (see the bottom-right plot.)
    %We compare the baseline network having a
    %content-based only attention mechanism (top row) with a hybrid
    %attention mechanism that uses convolutional features (bottom
    %row). 
  }
  \label{fig:forced_ali_corrected}

  \vspace{-4mm}
\end{figure}

\subsection{Forced Alignment of Long Utterances}

The good performance of the baseline model led us to the question of how it
distinguishes between repetitions of similar phoneme sequences and how reliably
it decodes longer sequences with more repetitions. We created two datasets of
long utterances; one by repeating each test utterance, and the other by
concatenating randomly chosen utterances. In both cases, the waveforms were
cross-faded with a 0.05s silence inserted as the ``pau'' phone. We concatenated
up to $15$ utterances.

First, we checked the forced alignment with these longer utterances by forcing
the generator to emit the correct phonemes. Each alignment was considered
correct if 90\% of the alignment weight lies inside the ground-truth phoneme
window extended by 20 frames on each side. Under this definition, all
phones but the
$\left<\text{eos}\right>$ shown in Fig.~\ref{fig:ali_baseline} are properly
aligned.

The first column of Fig.~\ref{fig:forced_ali_corrected} shows the number of
correctly aligned frames w.r.t. the utterance length (in frames) for some of the
considered models. One can see that the baseline model was able to decode
sequences up to about 120 phones when a single utterance was repeated, and up to
about 150 phones when different utterances were concatenated. Even when it
failed, it correctly aligned about 50 phones. On the other hand, the model with
the hybrid attention mechanism with convolutional features was able to align
sequences up to 200 phones long. However, once it began to fail, the model was
not able to align almost all phones. The model with the smoothing behaved
similarly to the one with convolutional features only.

%Todo: plots for the appendix, or maybe show them here??
We examined failed alignments to understand these two different modes of
failure. Some of the examples are shown in the Supplementary Materials.

We found that the baseline model properly aligns about 40 first phones, then
makes a jump to the end of the recording and cycles over the last 10 phones.
This behavior suggests that it learned to track its approximate location in the
source sequence. However, the tracking capability is limited to the lengths
observed during training. Once the tracker saturates, it jumps to the end of the
recording.
%(the distance to the end was supposedly learned by the reverse RNN
%in the Encoder). 

In contrast, when the location-aware network failed it just stopped aligning --
no particular frames were selected for each phone.  We attribute this behavior
to the issue of noisy glimpse discussed in Sec.~\ref{sec:sharpening}. With a
long utterance there are many 
%hundred 
irrelevant frames negatively affecting the weight assigned to the correct
frames. In line with this conjecture, the location-aware network works slightly
better on the repetition of the same utterance, where all frames are somehow
relevant, than on the concatenation of different utterances, where each
misaligned frame is irrelevant. % to the currently decoded phone.

To gain more insight 
%about the ways the networks failed 
we applied the alignment sharpening schemes described in
Sec.~\ref{sec:sharpening}. In the remaining columns of
Fig.~\ref{fig:forced_ali_corrected}, we see that the sharpening
methods help the location-aware network to find proper alignments, while they
show little effect on the baseline network. 
%Spatially constraining the alignment
%to the vicinity of the previous one (i.e. windowing) 
The windowing technique helps both the baseline and location-aware networks,
with the location-aware network properly aligning nearly all sequences. 

During visual inspection, we noticed that in the middle of very long utterances
the baseline model was confused by repetitions of similar content within the
window, and that such confusions did not happen in the beginning. This supports
our conjecture above.
% DIMA: figure in Appendix would be so great

\begin{figure}[t]
  \centering
  \includegraphics[width=\textwidth]{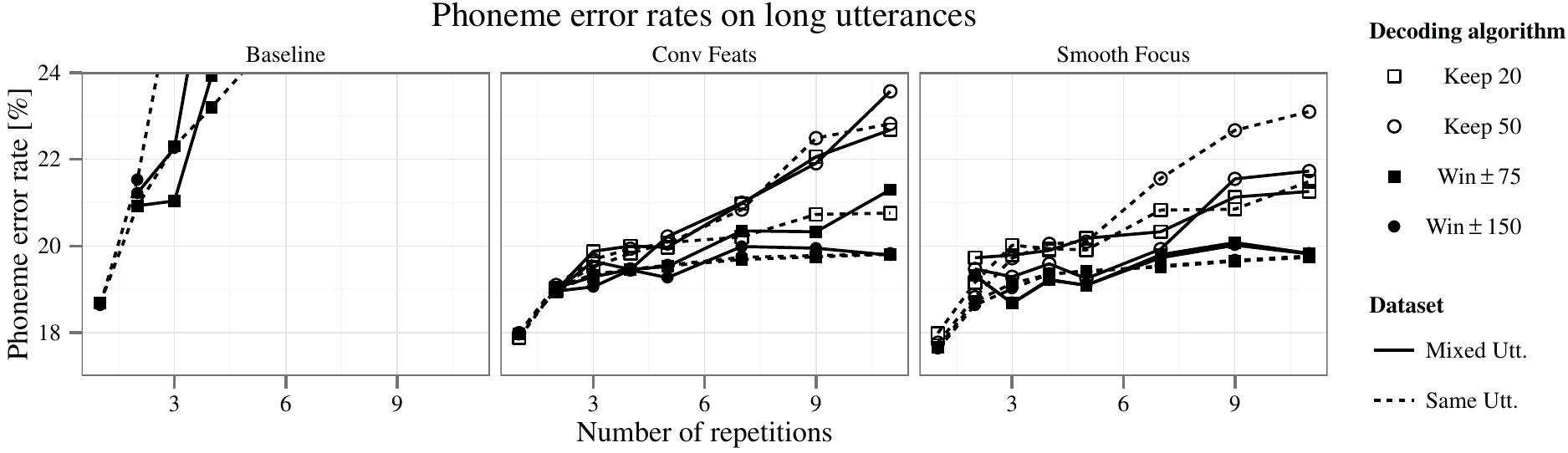}
  %\vspace{-.8cm}
  \caption[Phoneme error rates obtained on decoding long sequences.]{ 
      Phoneme error rates obtained on decoding long sequences. Each network was
      decoded with alignment sharpening techniques that produced proper forced
      alignments. The proposed ARSG's are clearly more robust to the length of
      the utterances than the baseline one is.
  }
  \label{fig:decoding_longs}

  \vspace{-4mm}
\end{figure}

\subsection{Decoding Long Utterances}

We evaluated the models on long sequences. Each model was decoded using the
alignment sharpening techniques that helped to obtain proper forced alignments.
The results are presented in Fig.~\ref{fig:decoding_longs}. The baseline model
fails to decode long utterances, even when a narrow window is used to constrain
the alignments it produces. The two other location-aware networks 
%with convolutional features added to the attention mechanism 
are able to decode utterances formed by concatenating up to 11 test utterances.
Better results were obtained with a wider window, presumably because it
resembles more the training conditions when at each step the attention mechanism
was seeing the whole input sequence.  With the wide window, both of the networks
scored about 20\% PER on the long utterances, indicating that the proposed
location-aware attention mechanism can scale to sequences much longer than those
in the training set with only minor modifications required at the decoding
stage.

\section{Conclusions}

We proposed and evaluated a novel end-to-end trainable speech recognition
architecture based on a hybrid attention mechanism which combines both content
and location information in order to select the next position in the input
sequence for decoding. One desirable property of the proposed model is that it
can recognize utterances much longer than the ones it was trained on. In the
future, we expect this model to be used to directly recognize text from
speech~\cite{hannun2014_deepspeech,graves_2014_towards}, in which case  it may
become important to incorporate a monolingual language model to the ARSG
architecture \cite{gulcehre_2015}.

This work has contributed two novel ideas for attention mechanisms: a better
normalization approach yielding smoother alignments and a generic principle for
extracting and using features from the previous alignments. 
%and using this features as additional low-level inputs to the
%attention mechanism. 
Both of these can potentially be applied beyond speech recognition. For
instance, the proposed attention can be used without
modification in neural Turing machines, or by using 2--D
convolution instead of 1--D, for improving image caption
generation~\cite{xu_show_2015}.

\subsubsection*{Acknowledgments}
All experiments were conducted using Theano
\cite{bergstra+al:2010-scipy,Bastien-Theano-2012}, PyLearn2
\cite{pylearn2_arxiv_2013}, and Blocks
\cite{vanmerrienboer_blocks_2015} libraries.

The authors would like to acknowledge the support of the following agencies for
research funding and computing support: National Science Center (Poland), 
NSERC, Calcul Qu\'{e}bec, Compute Canada,
the Canada Research Chairs and CIFAR. Bahdanau also thanks Planet
Intelligent Systems GmbH and Yandex.

{\small
\bibliographystyle{unsrt}
\bibliography{paper}

\begin{thebibliography}{10}

\bibitem{graves_generating_2013}
Alex Graves.
\newblock Generating sequences with recurrent neural networks.
\newblock {\em {arXiv}:1308.0850}, August 2013.

\bibitem{bahdanau_neural_2014}
Dzmitry Bahdanau, Kyunghyun Cho, and Yoshua Bengio.
\newblock Neural machine translation by jointly learning to align and
  translate.
\newblock {\em {arXiv}:1409.0473}, September 2014.

\bibitem{xu_show_2015}
Kelvin Xu, Jimmy Ba, Ryan Kiros, Kyunghyun Cho, Aaron Courville, Ruslan
  Salakhutdinov, Richard Zemel, and Yoshua Bengio.
\newblock Show, attend and tell: Neural image caption generation with visual
  attention.
\newblock {\em {arXiv}:1502.03044}, February 2015.

\bibitem{mnih_2014}
Volodymyr Mnih, Nicolas Heess, Alex Graves, et~al.
\newblock Recurrent models of visual attention.
\newblock In {\em Advances in Neural Information Processing Systems}, pages
  2204--2212, 2014.

\bibitem{chorowski_2014}
Jan Chorowski, Dzmitry Bahdanau, Kyunghyun Cho, and Yoshua Bengio.
\newblock End-to-end continuous speech recognition using attention-based
  recurrent {NN}: First results.
\newblock {\em {arXiv}:1412.1602 {[}cs, stat{]}}, December 2014.

\bibitem{graves_2014}
Alex Graves, Greg Wayne, and Ivo Danihelka.
\newblock Neural turing machines.
\newblock {\em ar{X}iv:{\tt 1410.5401}}, 2014.

\bibitem{weston_2014}
Jason Weston, Sumit Chopra, and Antoine Bordes.
\newblock Memory networks.
\newblock {\em ar{X}iv:{\tt 1410.3916}}, 2014.

\bibitem{gales_application_2007}
Mark Gales and Steve Young.
\newblock The application of hidden markov models in speech recognition.
\newblock {\em Found. Trends Signal Process.}, 1(3):195--304, January 2007.

\bibitem{hinton_deep_2012}
G.~Hinton, Li~Deng, Dong Yu, G.E. Dahl, A~Mohamed, N.~Jaitly, A~Senior,
  V.~Vanhoucke, P.~Nguyen, T.N. Sainath, and B.~Kingsbury.
\newblock Deep neural networks for acoustic modeling in speech recognition: The
  shared views of four research groups.
\newblock {\em {IEEE} Signal Processing Magazine}, 29(6):82--97, November 2012.

\bibitem{hannun2014_deepspeech}
Awni Hannun, Carl Case, Jared Casper, Bryan Catanzaro, Greg Diamos, Erich
  Elsen, Ryan Prenger, Sanjeev Satheesh, Shubho Sengupta, Adam Coates, et~al.
\newblock Deepspeech: Scaling up end-to-end speech recognition.
\newblock {\em arXiv preprint arXiv:1412.5567}, 2014.

\bibitem{hochreiter_1997}
S.~Hochreiter and J.~Schmidhuber.
\newblock Long short-term memory.
\newblock {\em Neural. Comput.}, 9(8):1735--1780, 1997.

\bibitem{cho_2014}
Kyunghyun Cho, Bart van Merrienboer, Caglar Gulcehre, Fethi Bougares, Holger
  Schwenk, and Yoshua Bengio.
\newblock Learning phrase representations using {RNN} encoder-decoder for
  statistical machine translation.
\newblock In {\em EMNLP 2014}, October 2014.
\newblock to appear.

\bibitem{graves_2006_connectionist}
Alex Graves, Santiago Fern{\'a}ndez, Faustino Gomez, and J{\"u}rgen
  Schmidhuber.
\newblock Connectionist temporal classification: Labelling unsegmented sequence
  data with recurrent neural networks.
\newblock In {\em ICML-06}, 2006.

\bibitem{graves_2012_sequence}
Alex Graves.
\newblock Sequence transduction with recurrent neural networks.
\newblock In {\em ICML-12}, 2012.

\bibitem{LeCun98-small}
Y.~{LeCun}, L.~Bottou, Y.~Bengio, and P.~Haffner.
\newblock Gradient based learning applied to document recognition.
\newblock {\em Proc. IEEE}, 1998.

\bibitem{graves_2013_timit}
Alex Graves, Abdel-rahman Mohamed, and Geoffrey Hinton.
\newblock Speech recognition with deep recurrent neural networks.
\newblock In {\em ICASSP 2013}, pages 6645--6649. IEEE, 2013.

\bibitem{graves_2014_towards}
Alex Graves and Navdeep Jaitly.
\newblock Towards end-to-end speech recognition with recurrent neural networks.
\newblock In {\em ICML-14}, pages 1764--1772, 2014.

\bibitem{sukhbaatar_2015}
Sainbayar Sukhbaatar, Arthur Szlam, Jason Weston, and Rob Fergus.
\newblock Weakly supervised memory networks.
\newblock {\em arXiv preprint arXiv:1503.08895}, 2015.

\bibitem{timit}
J.~S. Garofolo, L.~F. Lamel, W.~M. Fisher, J.~G. Fiscus, D.~S. Pallett, and
  N.~L. Dahlgren.
\newblock {DARPA} {TIMIT} acoustic phonetic continuous speech corpus, 1993.

\bibitem{povey_2011}
Daniel Povey, Arnab Ghoshal, Gilles Boulianne, Lukas Burget, Ondrej Glembek,
  Nagendra Goel, Mirko Hannemann, Petr Motlicek, Yanmin Qian, Petr Schwarz, and
  {others}.
\newblock The kaldi speech recognition toolkit.
\newblock In {\em Proc. {ASRU}}, pages 1--4, 2011.

\bibitem{zeiler_2012}
Matthew~D Zeiler.
\newblock {ADADELTA}: An adaptive learning rate method.
\newblock {\em ar{X}iv:{\tt 1212.5701}}, 2012.

\bibitem{graves_2011}
Alex Graves.
\newblock Practical variational inference for neural networks.
\newblock In J.~Shawe-Taylor, R.S. Zemel, P.L. Bartlett, F.~Pereira, and K.Q.
  Weinberger, editors, {\em Advances in Neural Information Processing Systems
  24}, pages 2348--2356. Curran Associates, Inc., 2011.

\bibitem{hinton_2012}
Geoffrey~E Hinton, Nitish Srivastava, Alex Krizhevsky, Ilya Sutskever, and
  Ruslan~R Salakhutdinov.
\newblock Improving neural networks by preventing co-adaptation of feature
  detectors.
\newblock {\em arXiv preprint arXiv:1207.0580}, 2012.

\bibitem{sutskever_sequence_2014}
Ilya Sutskever, Oriol Vinyals, and Quoc~V. Le.
\newblock Sequence to sequence learning with neural networks.
\newblock {\em {arXiv} preprint {arXiv}:1409.3215}, 2014.

\bibitem{toth_2014}
L{\'a}szl{\'o} T{\'o}th.
\newblock Combining time-and frequency-domain convolution in convolutional
  neural network-based phone recognition.
\newblock In {\em ICASSP 2014}, pages 190--194, 2014.

\bibitem{gulcehre_2015}
Caglar Gulcehre, Orhan Firat, Kelvin Xu, Kyunghyun Cho, Loic Barrault, Huei-Chi
  Lin, Fethi Bougares, Holger Schwenk, and Yoshua Bengio.
\newblock On using monolingual corpora in neural machine translation.
\newblock {\em arXiv preprint arXiv:1503.03535}, 2015.

\bibitem{bergstra+al:2010-scipy}
James Bergstra, Olivier Breuleux, Fr{\'{e}}d{\'{e}}ric Bastien, Pascal Lamblin,
  Razvan Pascanu, Guillaume Desjardins, Joseph Turian, David Warde-Farley, and
  Yoshua Bengio.
\newblock Theano: a {CPU} and {GPU} math expression compiler.
\newblock In {\em Proceedings of the Python for Scientific Computing Conference
  ({SciPy})}, June 2010.
\newblock Oral Presentation.

\bibitem{Bastien-Theano-2012}
Fr{\'{e}}d{\'{e}}ric Bastien, Pascal Lamblin, Razvan Pascanu, James Bergstra,
  Ian~J. Goodfellow, Arnaud Bergeron, Nicolas Bouchard, and Yoshua Bengio.
\newblock Theano: new features and speed improvements.
\newblock Deep Learning and Unsupervised Feature Learning NIPS 2012 Workshop,
  2012.

\bibitem{pylearn2_arxiv_2013}
Ian~J. Goodfellow, David Warde-Farley, Pascal Lamblin, Vincent Dumoulin, Mehdi
  Mirza, Razvan Pascanu, James Bergstra, Fr{\'{e}}d{\'{e}}ric Bastien, and
  Yoshua Bengio.
\newblock Pylearn2: a machine learning research library.
\newblock {\em arXiv preprint arXiv:1308.4214}, 2013.

\bibitem{vanmerrienboer_blocks_2015}
Bart van Merri{\"e}nboer, Dzmitry Bahdanau, Vincent Dumoulin, Dmitriy Serdyuk,
  David Warde-Farley, Jan Chorowski, and Yoshua Bengio.
\newblock Blocks and fuel: Frameworks for deep learning.
\newblock {\em {arXiv}:1506.00619 {[}cs, stat{]}}, June 2015.

\end{thebibliography}
}

\clearpage
\appendix 
\section{Additional Figures}

\begin{figure}[h]
  \centering
  Baseline
  \includegraphics[width=\textwidth]{michael_baseline}

  Convolutional Features
  \includegraphics[width=\textwidth]{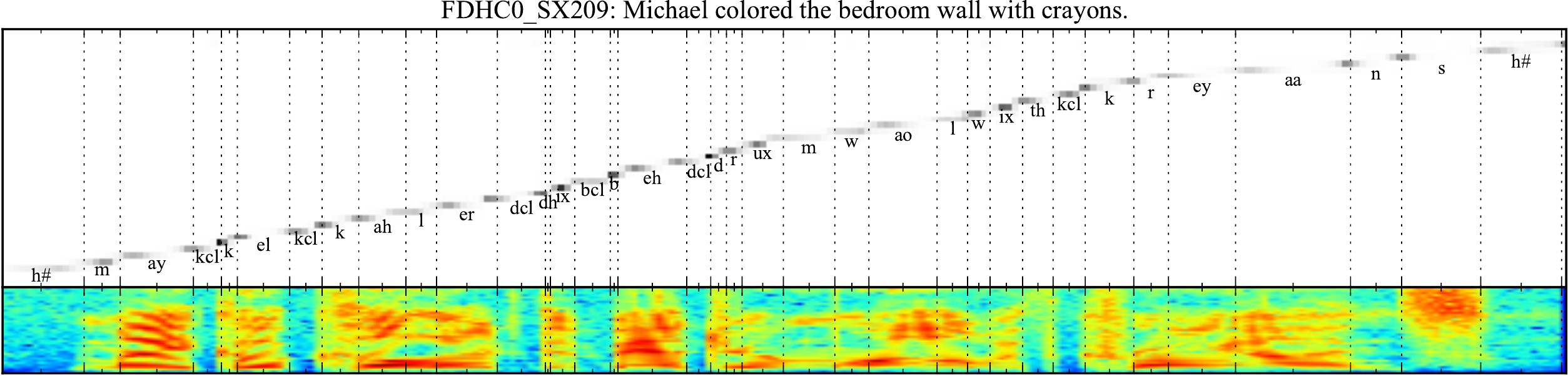}
  Smooth Focus
  \includegraphics[width=\textwidth]{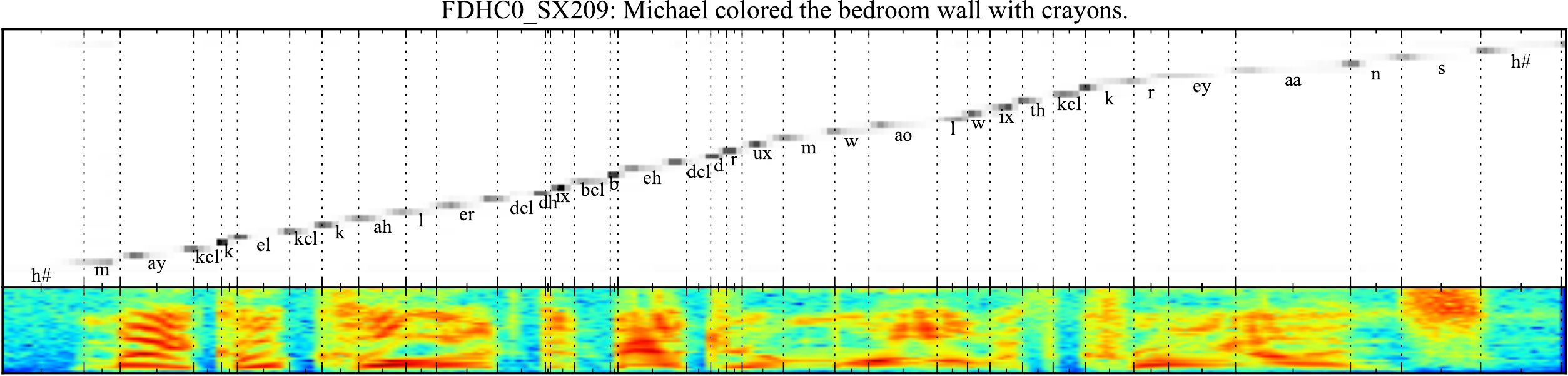}
  % \vspace{-.8cm}
  \caption{Alignments
      produced by evaluated models on the FDHC0\_SX209 test utterance. The vertical bars
      indicate ground truth phone location from TIMIT. Each
      row of the upper image indicates frames selected by
      the attention mechanism to emit a phone symbol. Compare with
      Figure 3. in the main text.
  }  

  \vspace{-4mm}
\end{figure}

\begin{figure}[h]
  \centering
  Baseline
  \includegraphics[width=\textwidth]{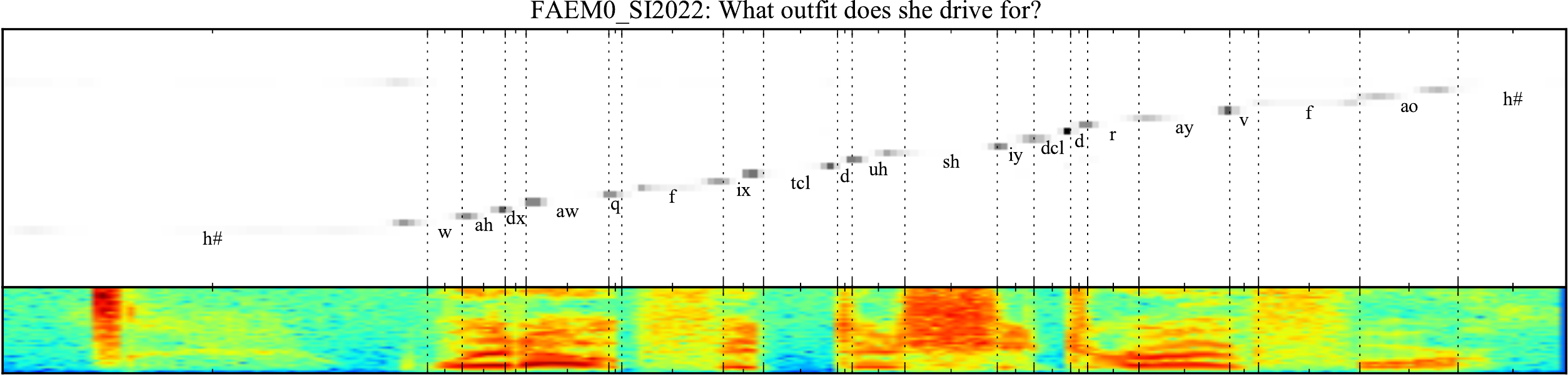}

  Convolutional Features
  \includegraphics[width=\textwidth]{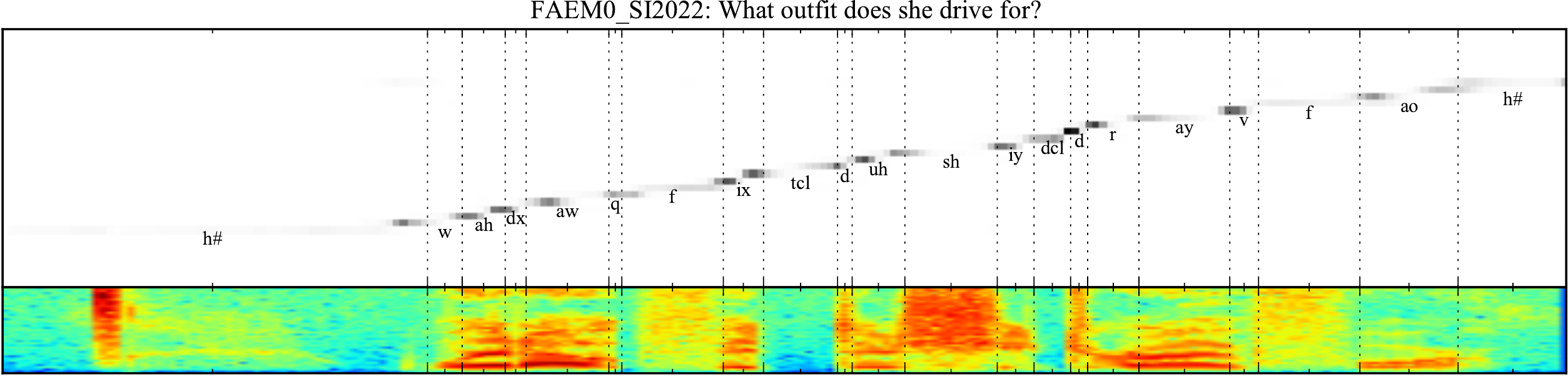}
  Smooth Focus
  \includegraphics[width=\textwidth]{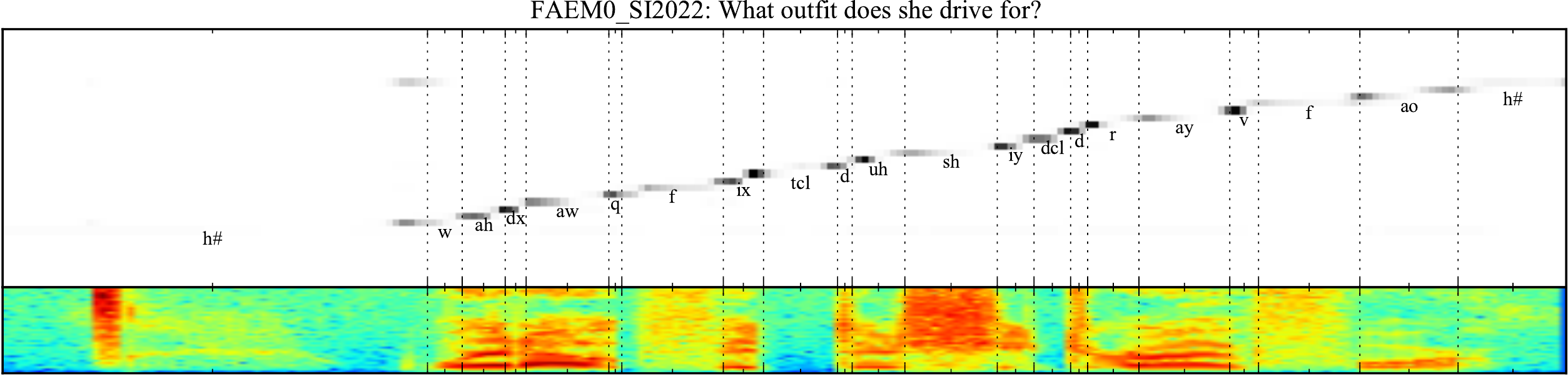}
  % \vspace{-.8cm}
  \caption{Alignments
      produced by evaluated models on the FAEM0\_SI2022 train utterance. The vertical bars
      indicate ground truth phone location from TIMIT. Each
      row of the upper image indicates frames selected by
      the attention mechanism to emit a phone symbol. Compare with
      Figure 3. in the main text.
  }  

  \vspace{-4mm}
\end{figure}

\begin{figure}[h]
  \centering
  \includegraphics[width=\textwidth]{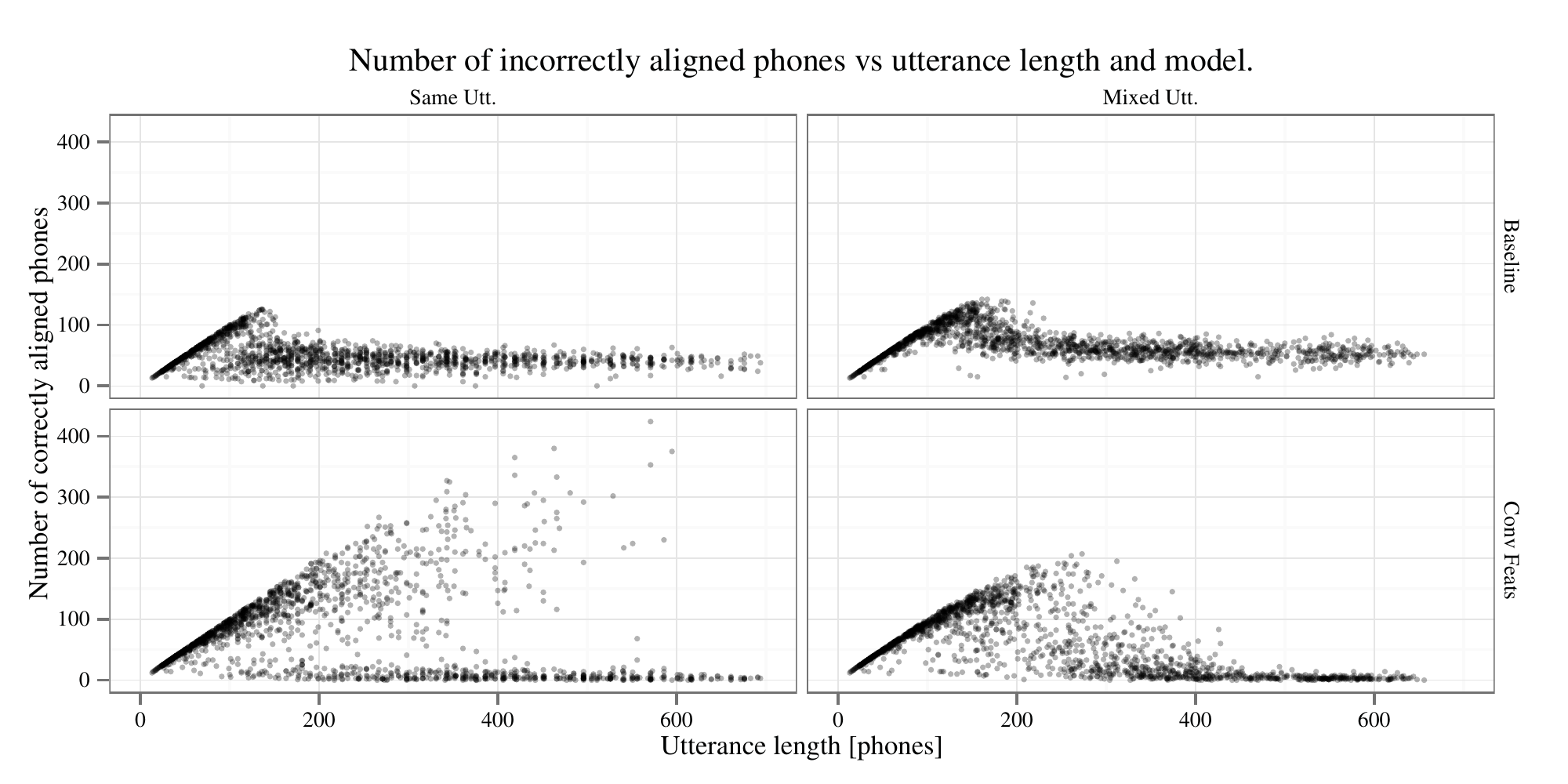}
  \caption[Results of force-aligning of long utterances.]{Close-up on
    the two failure modes of ARSG. Results of
    force-aligning concatenated TIMIT utterances. Each dot represents
    a single utterance. The left panels show results for
    concatenations of the same utterance. The right panels show
    results for concatenations of randomly chosen utterances. We
    compare the baseline network having a content-based only attention
    mechanism (top row) with a hybrid attention mechanism that uses
    convolutional features (bottom row). While neither model is able
    to properly align long sequences, they fail in different ways: the
    baseline network always aligns about 50 phones, while the
    location-aware network fails to align any phone. Compare with
    Figure 4 form the main paper.}
  %\label{fig:forced_ali_fail}
\end{figure}

\begin{sidewaysfigure}[h]
  \includegraphics[width=\textwidth]{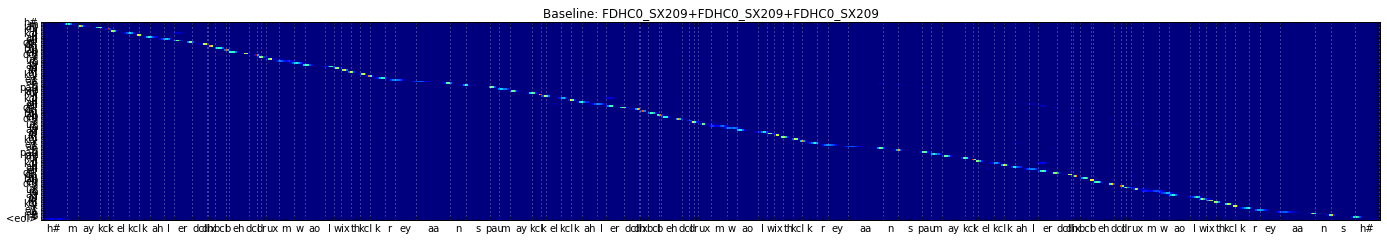}
  \includegraphics[width=\textwidth]{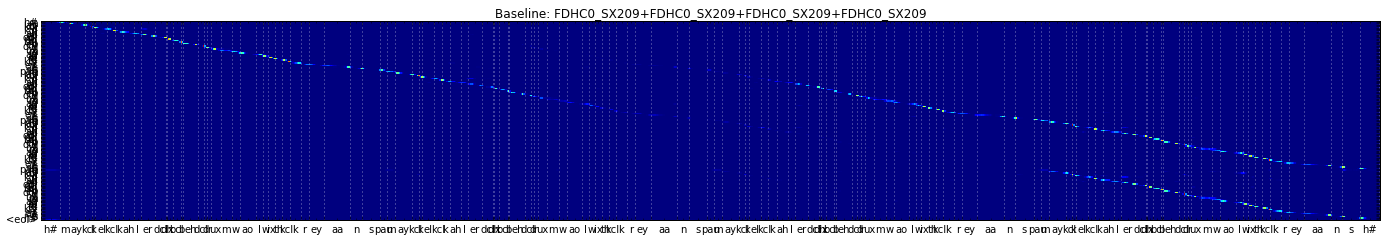}
  \includegraphics[width=\textwidth]{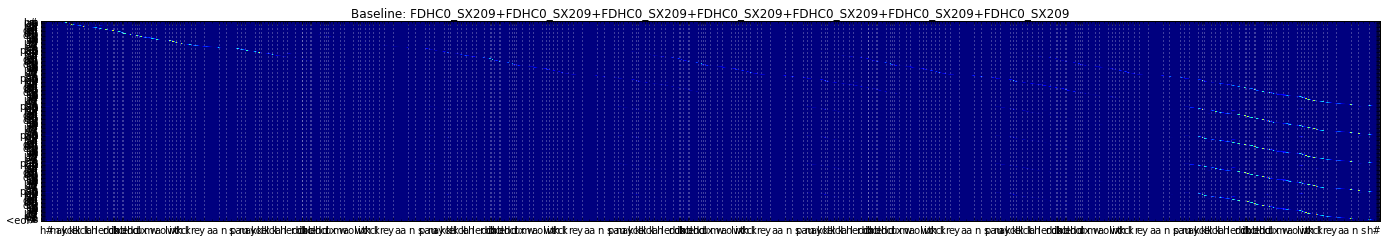}
  \includegraphics[width=\textwidth]{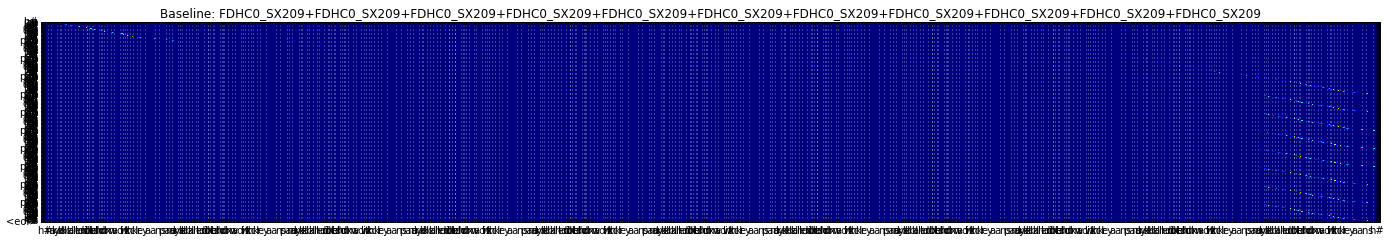}
  \caption{The baseline network fails to align more than 3 repetitions
    of FDHC0\_SX209.}
\end{sidewaysfigure}

\begin{sidewaysfigure}[h]
  \includegraphics[width=\textwidth]{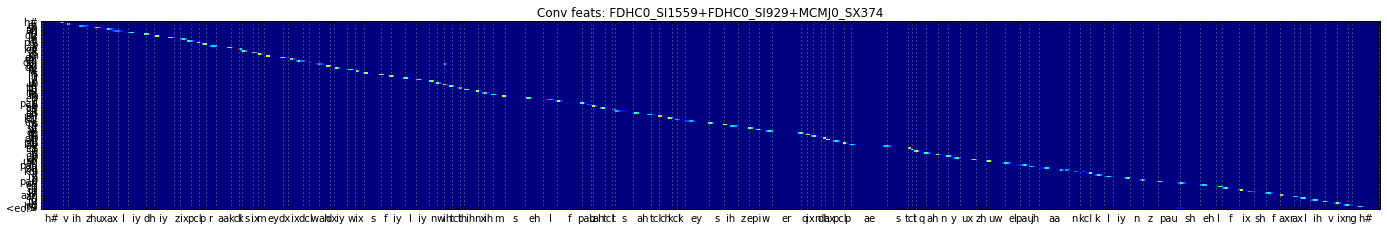}
  \includegraphics[width=\textwidth]{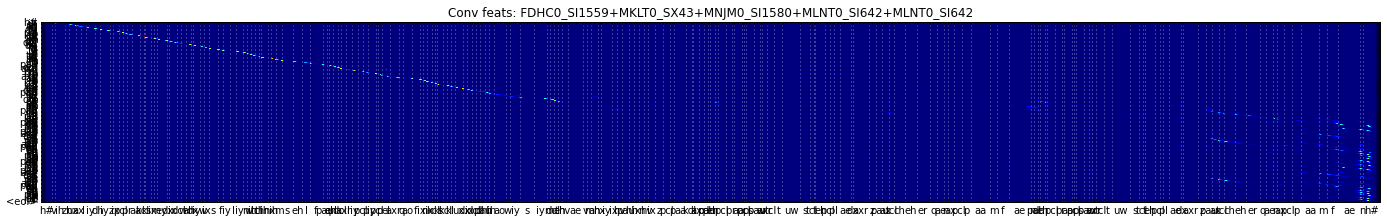}
  \caption{The baseline network aligns a concatenation of 3 different
    utterances, but fails to align 5.}
\end{sidewaysfigure}

\begin{sidewaysfigure}[h]
  \includegraphics[width=\textwidth]{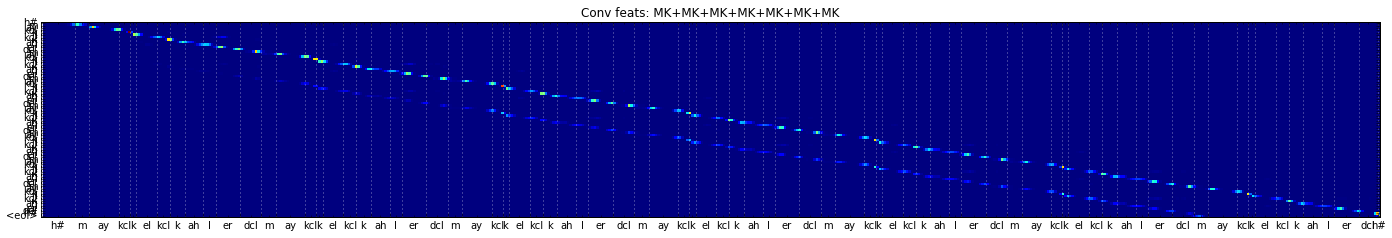}
  \caption{Forced alignment of 7 repetitions of the phrase ``Michael
    colored'' performed with the baseline model with windowing enabled
  (the alignment was constrained to $\pm 75$ frames from the expected
  position of the generator at the last step. The window is wider than
the pattern and the net confuses similar content. Strangely, the first
two repetitions are aligned without any confusion with subsequent ones --
the network starts to confound phoneme location only starting from the
third repetition (as seen by the parallel strand of alignment which
starts when the network starts to emit the phrase for the third time).
}
\end{sidewaysfigure}

\begin{sidewaysfigure}[h]
  \includegraphics[width=\textwidth]{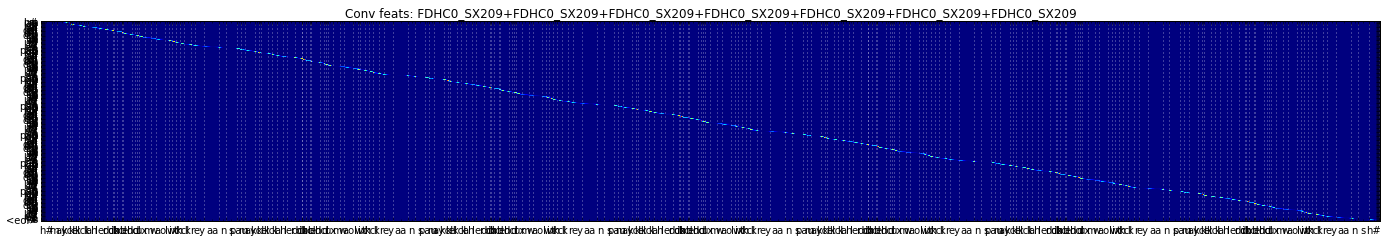}
  \includegraphics[width=\textwidth]{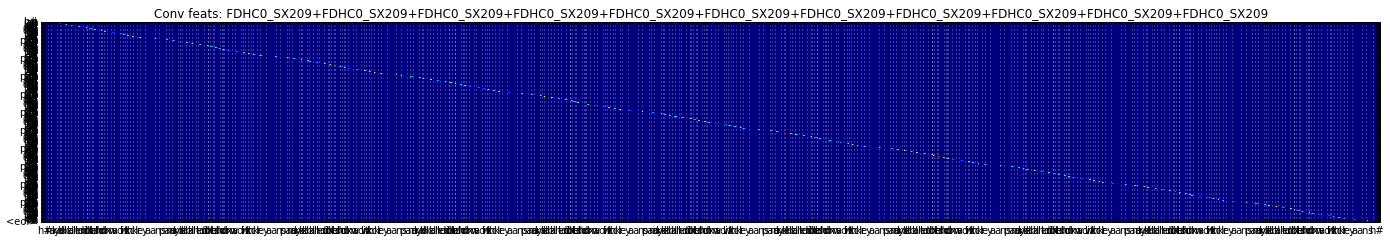}
  \includegraphics[width=\textwidth]{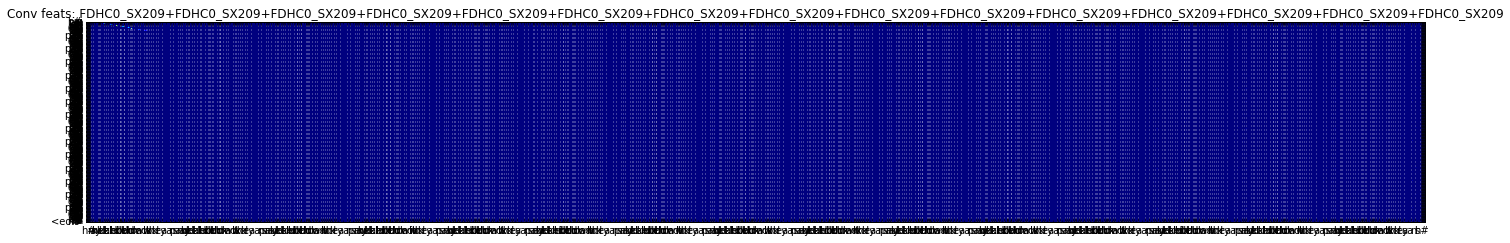}
  \caption{The location-aware network correctly aligns 7 and 11
    repetitions of FDHC0\_SX209, butfails to align 15 repetitions
    of FDHC0\_SX209.}
\end{sidewaysfigure}

\begin{sidewaysfigure}[h]
  \includegraphics[width=\textwidth]{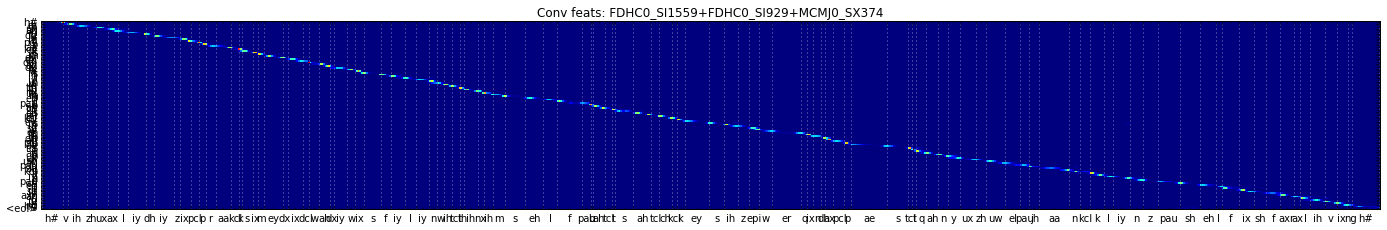}
  \includegraphics[width=\textwidth]{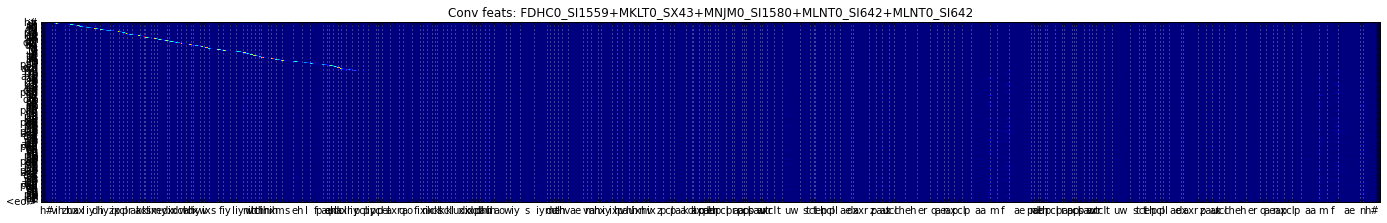}
  \caption{The location-aware network aligns a concatenation of 3 different
    utterances, but fails to align 5.}
\end{sidewaysfigure}

\clearpage
\section{Detailed results of experiments}

\begin{table}[h]
  \centering
  \caption{Phoneme error rates while decoding with various
    modifications. Compare with Figure 5 from the main paper.}
%narrow one without dev errs
% \begin{tabular}{lrrrrrrrr}
% algorithm &   Plain &         &  Keep 1 & Keep 10 & Keep 50 &
% $\beta=2$ & Win. $\pm 75$ & Win. $\pm 150$ \\
% dataset &     dev &    test &    test &    test &    test &    test &       test &        test \\
% Baseline     & $15.9\%$ & $18.7\%$ & $20.2\%$ & $18.7\%$ & $18.7\%$ & $18.9\%$ &    $18.7\%$ &     $18.6\%$ \\
% Conv Feats   & $16.1\%$ & $18.0\%$ & $22.3\%$ & $17.9\%$ & $18.0\%$ & $18.7\%$ &    $18.0\%$ &     $18.0\%$ \\
% Smooth Focus & $15.8\%$ & $17.6\%$ & $24.7\%$ & $18.7\%$ & $17.8\%$ & $18.4\%$ &    $17.7\%$ &     $17.6\%$ \\
% \end{tabular}
\setlength\tabcolsep{5pt}
\begin{tabular}{rl|c|c|c|c|c|c|c}
             &         &   Plain  &  Keep 1  & Keep 10  & Keep 50  &
             $\beta=2$ & Win. $\pm 75$ & Win. $\pm 150$ \\ \hline \hline
\multirow{2}{*}{Baseline} &     dev & $15.9\%$ & $17.6\%$ & $15.9\%$ & $15.9\%$ &   $16.1\%$ &       $15.9\%$ &        $15.9\%$ \\
             &    test & $18.7\%$ & $20.2\%$ & $18.7\%$ & $18.7\%$ &
             $18.9\%$ &       $18.7\%$ &        $18.6\%$ \\ \hline
\multirow{2}{*}{Conv Feats} &     dev & $16.1\%$ & $19.4\%$ & $16.2\%$ & $16.1\%$ &   $16.7\%$ &       $16.0\%$ &        $16.1\%$ \\
             &    test & $18.0\%$ & $22.3\%$ & $17.9\%$ & $18.0\%$ &
             $18.7\%$ &       $18.0\%$ &        $18.0\%$ \\ \hline
\multirow{2}{*}{Smooth Focus} &     dev & $15.8\%$ & $21.6\%$ & $16.5\%$ & $16.1\%$ &   $16.2\%$ &       $16.2\%$ &        $16.0\%$ \\
             &    test & $17.6\%$ & $24.7\%$ & $18.7\%$ & $17.8\%$ &   $18.4\%$ &       $17.7\%$ &        $17.6\%$ \\
\end{tabular}
  \label{tab:decoding_singles}
\end{table}

%{\small
%\bibliographystyle{unsrt}
%\bibliography{paper}
%}

\end{document}